%% file: root.tex
\documentclass[letterpaper, 10 pt, conference]{ieeeconf}  

\IEEEoverridecommandlockouts                              

\usepackage{subfiles}
\usepackage{amsmath}
\usepackage{amssymb}
\usepackage{mathtools}
\usepackage{graphicx}
\usepackage{float}
\usepackage{subcaption}
\usepackage{epstopdf}
\usepackage[tableposition=top]{caption}
\usepackage{array}
\usepackage{url}
\usepackage{algorithm}
\usepackage{algpseudocode}
\usepackage{diagbox}
\usepackage{multirow}
\usepackage[table]{xcolor}
\usepackage[framemethod=TikZ]{mdframed}
\usepackage[procnames]{listings}
\usepackage[hidelinks]{hyperref}
\usepackage{bm}

\mdfdefinestyle{frameStyle}{%
    roundcorner=5pt,
    backgroundcolor=white}

\algtext*{EndIf}
\algtext*{EndFor}
\algtext*{EndWhile}
\algtext*{EndFunction}

\newcommand{\assign}[0]{$\leftarrow$ }
\newcommand{\algto}[0]{\textbf{to} }

\renewcommand{\vec}[1]{\bm{#1}}

\title{\LARGE \bf
Learning Human Body Motions from Skeleton-Based Observations\\for Robot-Assisted Therapy
}

\author{Natalia Quiroga$^{\dagger\mathsection}$, Alex Mitrevski$^{\dagger\mathsection}$, Paul G. Pl{\"o}ger$^{\dagger}$
\thanks{$^{\dagger}$The authors are with the Autonomous Systems Group, Department of Computer Science, Hochschule Bonn-Rhein-Sieg, Sankt Augustin, Germany
        {\tt\scriptsize <aleksandar.mitrevski, paul.ploeger>@h-brs.de, natalia.quiroga@smail.inf.h-brs.de}} %
\thanks{$^{\mathsection}$Corresponding author} %
}

\begin{document}

\maketitle
\thispagestyle{empty}
\pagestyle{empty}


\begin{abstract}
    Robots applied in therapeutic scenarios, for instance in the therapy of individuals with Autism Spectrum Disorder, are sometimes used for imitation learning activities in which a person needs to repeat motions by the robot.
    To simplify the task of incorporating new types of motions that a robot can perform, it is desirable that the robot has the ability to learn motions by observing demonstrations from a human, such as a therapist.
    In this paper, we investigate an approach for acquiring motions from skeleton observations of a human, which are collected by a robot-centric RGB-D camera.
    Given a sequence of observations of various joints, the joint positions are mapped to match the configuration of a robot before being executed by a PID position controller.
    We evaluate the method, in particular the reproduction error, by performing a study with QTrobot in which the robot acquired different upper-body dance moves from multiple participants.
    The results indicate the method's overall feasibility, but also indicate that the reproduction quality is affected by noise in the skeleton observations.
\end{abstract}

\subfile{subfiles/intro}

\subfile{subfiles/related_work}

\subfile{subfiles/imitation}

\subfile{subfiles/experiments}

\subfile{subfiles/conclusions}

\addtolength{\textheight}{-12cm}   


\section*{ACKNOWLEDGMENT}

    This work is conducted in the context of the MigrAVE project, which is funded by the German Ministry of Education and Research (BMBF).
    We hereby thank our partners, M{\"u}nster University of Applied Sciences (FHM) and the RFH - University of Applied Sciences, Cologne.
    The research in the MigrAVE project has received an ethics approval by the German Psychological Society (DGPs).


\bibliographystyle{IEEEtran}
\bibliography{references}

\end{document}

%% file: subfiles/intro.tex
    \section{INTRODUCTION}
    \label{sec:introduction}

    In robot-assisted therapy scenarios, it is important to have robots that can be easily adapted to the needs of the therapy.
    In this context, adaptation can include the reconfiguration of robot programs used in the therapy, the development of new programs without involving a robotics expert, or the personalisation of a robot's behaviour to a specific user \cite{cabibihan2013, ravichandar2020}.
    Particularly in the context of rehabilitation or children-centered therapy, it may be necessary to include different types of motions that can be included in exercises performed by the patient.
    For instance, in the treatment of children with Autism Spectrum Disorder (ASD), some part of the therapy may be focused on teaching a child useful everyday motion patterns, such as brushing teeth or holding a knife for cutting \cite{iacono2011, Prabha2021}.
    To make it possible to include different types of motions in therapy sessions without the need for pre-programming them, it is desirable that a robot is able to learn such motions as required, potentially by observing the therapists themselves \cite{Fadli2018, Hussein2018}.

    A widely used technique that makes it possible to acquire motions without the need for explicit robot programming is learning from demonstration \cite{ravichandar2020, argall2009}.
    Demonstration-based learning allows a robot to observe a person performing some motion, or an action in general, so that it can then reproduce it accordingly.
    There are two main challenges that are present in this context.
    First of all, given that a robot and the demonstrating person are likely to have different embodiments, demonstrated motions need to be mapped to the robot's configuration space so that they can be executed on the robot's platform \cite{Koenemann_2014, Koenemann2012}.
    Related to this, a demonstration can be observed using different sensory modalities, such that the modality determines the complexity of the demonstration setup and the difficulty of converting demonstrations to executable robot commands \cite{argall2009, Breazeal2002}.
    For instance, using dedicated motion-capture sensors on the person's joints may simplify the data recording, but increases the complexity of the setup; on the other hand, using kinaesthetic teaching simplifies the demonstration mapping problem, but requires the robot platform to support this type of demonstration \cite{Botev2020}.
    A considerably simpler and natural way of demonstration is one in which the person performs the demonstrated activity without specialised equipment; in this case, the robot should visually extract the demonstration and map it to its own embodiment for subsequent execution.

    In this paper, we focus on the problem of learning upper-body motions demonstrated by a person and reproducing those on a robot, with the motivation of incorporating such motions in the therapy of children with ASD.
    We learn motions by observing a person in 3D skeleton data collected by a robot; the motions are then converted to match the robot's body \cite{vandeperre2015, Chen_2016} and are executed on the robot using a position controller.
    We perform preliminary experiments of the presented method with QTrobot \cite{luxai2017} in which various motions were demonstrated by 20 participants.
    The results demonstrate the feasibility of learning actions from skeleton-based data, but the accuracy of the motion reproduction is affected by noise in the skeletons and thus requires noisy demonstration frames to be removed from the learning data.

%% file: subfiles/related_work.tex
    \section{RELATED WORK}
    \label{sec:related_work}

	As mentioned above, there are different ways in which imitation learning can be performed.
    In \cite{Riley2003}, full-body imitation is simulated for 32 degrees of freedom (DOF) and real replication is performed for 7 DOF, such that external landmarks are placed on the person for motion tracking.
    Similarly, in \cite{Koenemann_2014, Koenemann2012}, the robot has to imitate certain paths and replicate the movements of a human; for this, the user has to wear external equipment on the body.
    In this work, we perform motion tracking by recording skeleton data from an RGB-D camera as in \cite{Fadli2018, Koenemann2012, Chen_2016, Assad_Uz_Zaman_2020, Sripada_2018, Liu2015, tahara2022}.

	One important issue when doing imitation learning is that, in most cases, the number of DOF of a robot differs from that of a person; for this reason, the observed human trajectories have to be converted to a simplified configuration so that joint trajectories on the robot can be calculated.
    In \cite{vandeperre2015}, a model-based imitation learning method for robots with different configurations is presented and tested for imitating six postures that represent emotional expressions.
    Similarly,  a system capable to imitate whole-body motion using a Kalman filter is presented in \cite{Montecillo2010}, where human motion marker points and a normalised skeleton are used to match the DOF between a human and a robot; however, the quality of motion transfer was not measured and the method is shown to work at low speeds with a limited number of motions.
    In this paper, we also use a model-based method similar to \cite{vandeperre2015} and skeleton data as in \cite{Montecillo2010} for performing human-robot configuration mapping, but we record 3D positions of skeleton joints, which we subsequently use to calculate the angular positions of the joints of interest.

	In the context of robot manipulation and imitation learning, an inverse kinematics (IK) solution is required for finding a desired joint configuration that does not lead to self-collisions.
    According to \cite{Li_2021}, the numerical solution of solving the IK is time-consuming and computationally expensive, such that a hybrid solution that combines a mathematical model and a neural network is presented.
    Some heuristic methods for solving the IK problem are model-based \cite{aristidou2011, rokbani2015, F_dor_2003} or neural network-based \cite{Csiszar_2017}; however, these methods are not always implemented in a real robot or may not be applicable for real-time robot execution.
    To identify self-collisions, we use a method inspired by \cite{Corrales2011}, namely we calculate the expected end effector position of the robot based on the measured joint angles; if a self-collision with respect to a known model of the robot is identified, the position component that leads to the collision is mapped to the closest point outside the robot's model so that the collision can be avoided.
    It should, however, be mentioned that, in the interest of space, the self-collision is not evaluated in the experiments presented in this paper.

%% file: subfiles/imitation.tex
    \section{DEMONSTRATION-BASED LEARNING FROM SKELETON DATA}
    \label{sec:learning}

    The objective of this paper is to develop a method that makes it possible to acquire robot motions by observing a human from skeleton data.
    We aim for a procedure that allows one-shot motion learning, namely it should be possible to learn a desired motion from a single demonstration.
    The concrete approach we discuss is applied to the social robot QTrobot and consists of (i) recording skeleton data of a human demonstration as a sequence of joint positions, (ii) mapping the sequence of positions to the robot's joints, and (iii) executing the sequence on the robot.
    In this section, we discuss the representation of human demonstrations and their mapping to the robot's body, as well as the execution of demonstrated motions on the robot.

    \subsection{Robot Description}
    \label{sec:learning:robot}

    In this paper, we use QTrobot \cite{luxai2017} as a robot platform, which is a small table-based social robot for therapeutic applications, particularly for children with ASD.
    \begin{figure}[tp]
        \centering
        \includegraphics[width=8cm]{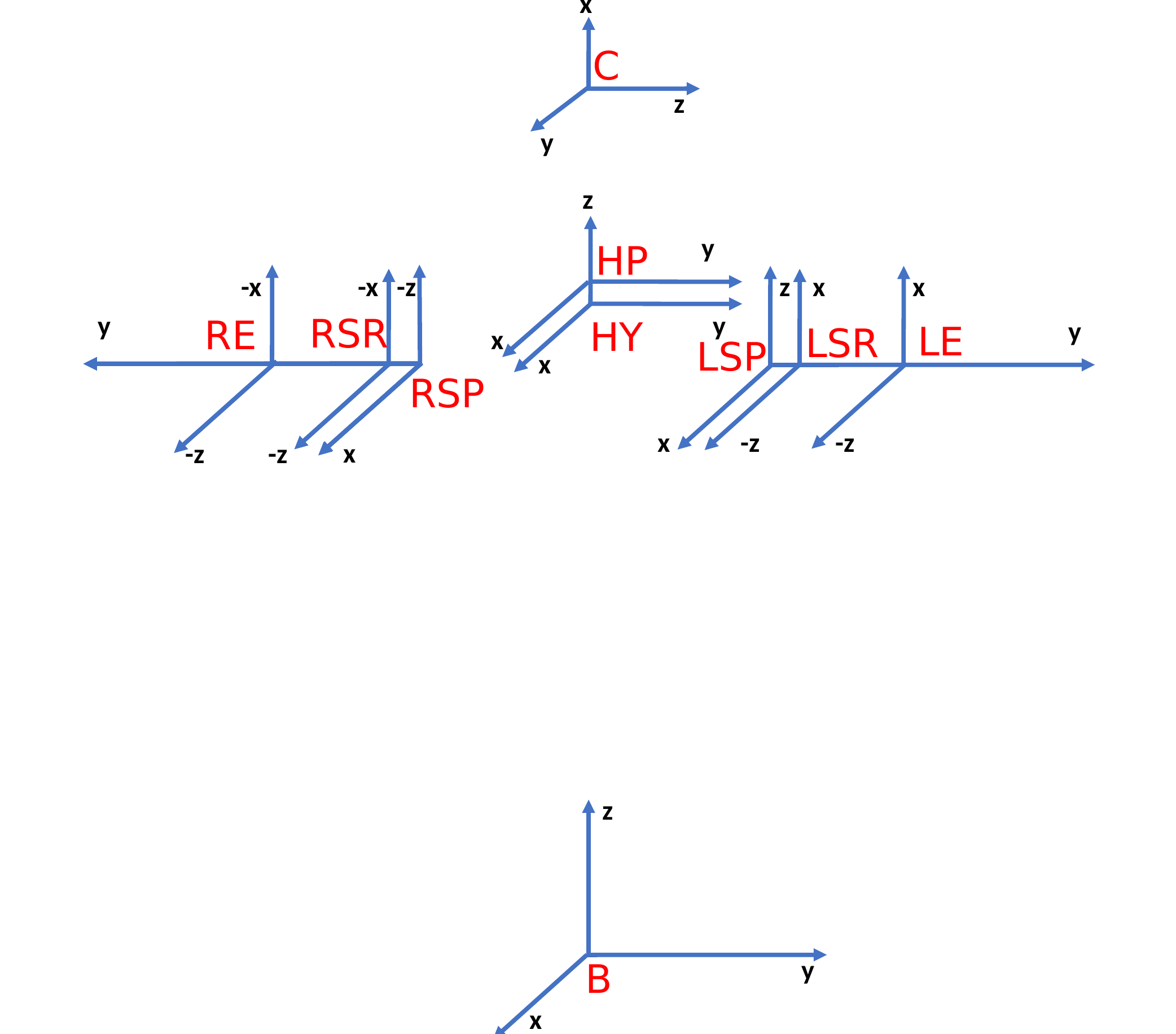}
        \caption{QTrobot's degrees of freedom (adapted from \cite{qtrobot_safety_manual})}
        \label{fig:qtrobot_configuration}
    \end{figure}
    The robot has a total of 12 degrees of freedom (DOF) in the head and arms, as shown in Fig. \ref {fig:qtrobot_configuration}; four of these are mechanical DOF, while eight are controllable, namely they have a motor and can be programmed.
    On the other hand, a reconstructed human skeleton has 24 joints taking into account the torso and legs.
    We focus on upper-body movements in this paper due to the fact that the robot's lower-body cannot be moved; thus, of the 24 human joints, seven are useful for the method presented in this paper, namely the head, neck, collar, left and right shoulders, elbows and wrists.

    \subsection{Demonstration Representation}
    \label{sec:learning:demonstration}

    We represent a demonstration of an action $a$ as a sequence $S_a$ of joint position measurements at multiple time points $t, 1 \leq t \leq T$:
    \begin{equation}
        S^H_a = \left( P^H_1, \hdots, P^H_T \right)
    \end{equation}
    where $P^H_t$ is a matrix that contains the positions of all $m$ recorded joints at time $t$:
    \begin{equation}
        P^H_t = \begin{pmatrix}
            \vec{p}^1_t & \hdots & \vec{p}^m_t
        \end{pmatrix} =
        \begin{pmatrix}
            x^1_t &        & x^m_t \\
            y^1_t & \hdots & y^m_t \\
            z^1_t &        & z^m_t
        \end{pmatrix}
    \end{equation}

    The position measurements in $P^H_t$ are used to calculate the angular positions $\theta^H_{t_j}, 1 \leq j \leq n$ of $n \leq m$ joints.
    The angles of all $n$ joints at time $t$ are thus represented as
    \begin{equation}
        \vec{\theta}^H_t = \left( \theta^H_{t_1} \hdots \theta^H_{t_n} \right)^T
    \end{equation}
    This results in an alternative representation $\Theta^H_a$ of a demonstration of $a$ in terms of the joint angles, which has the form
    \begin{equation}
        \Theta^H_a = \begin{pmatrix}
            \vec{\theta}^H_1, \hdots, \vec{\theta}^H_T
        \end{pmatrix} =
        \begin{pmatrix}
            \theta^H_{1_1} &        & \theta^H_{T_1} \\
            \vdots         & \hdots & \vdots \\
            \theta^H_{1_n} &        & \theta^H_{T_n} \\
        \end{pmatrix}
    \end{equation}

    The joint angles $\theta^H_{t,j}$ are calculated based on the human links, namely vectors that connect two joints whose positions are measured.
    In particular, given two links $\vec{l}_1$ and $\vec{l}_2$, the angle of joint $j$ is calculated based on the dot product:
    \begin{equation}
        \theta^H_j = \cos^{-1}\left(\frac{\vec{l}_1 \cdot \vec{l}_2}{\lVert \vec{l}_1 \rVert \lVert \vec{l}_2 \rVert}\right)
    \end{equation}

\begin{figure}[tp]
        \centering
        \includegraphics[width=0.9\linewidth]{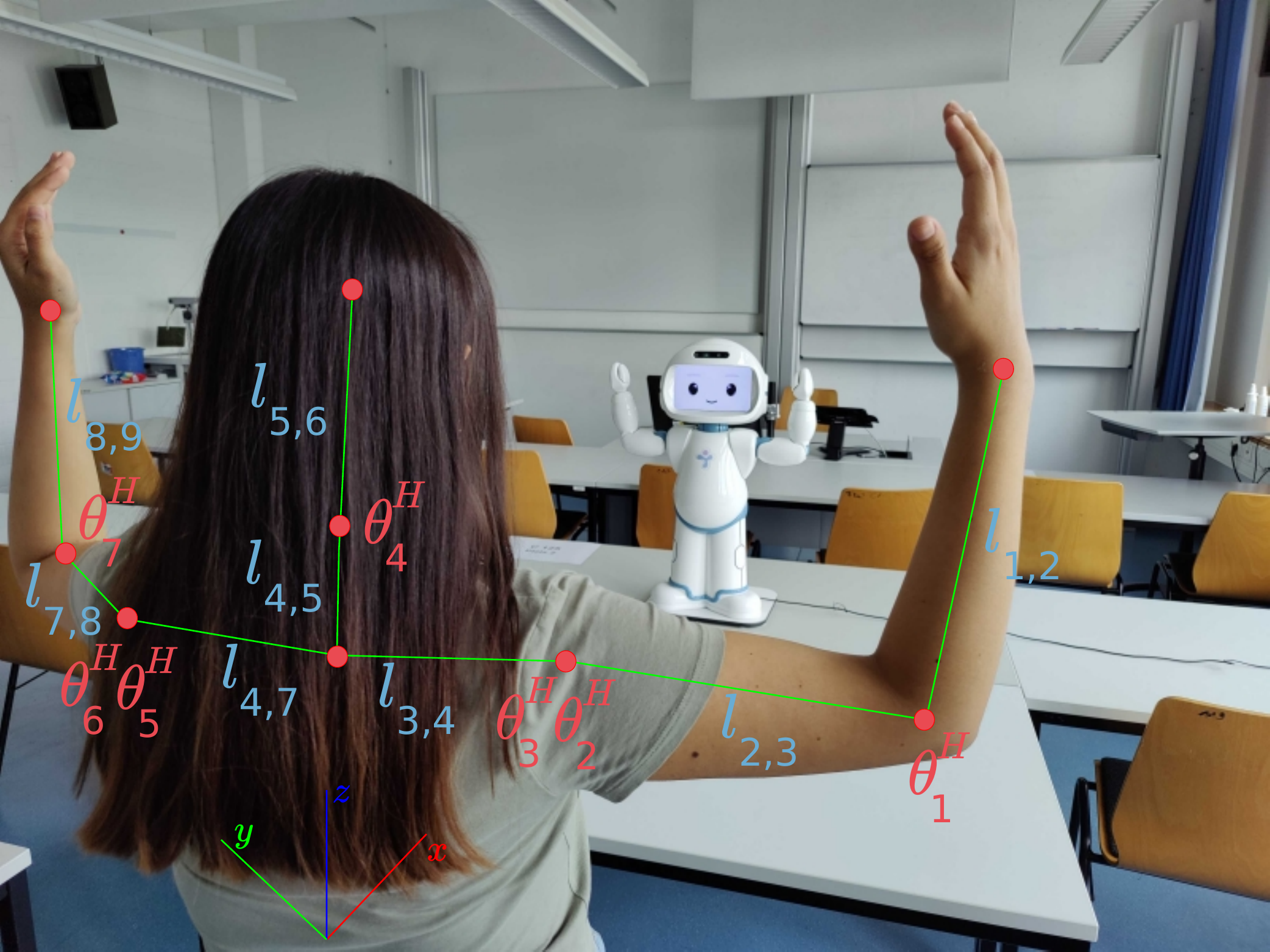}
        \caption{Illustration of the skeleton annotation}
        \label{fig:skeleton_annotation}
    \end{figure}

    The joints of interest in the human skeleton data as well as the links that are used to calculate the joint angles are listed in Tab. \ref{tab:links_for_joint_angle_calculations}.
    \begin{table}[t]
        \centering
		\caption{Links used to calculate the joint angles $\theta^H_{t_j}$}
        \label{tab:links_for_joint_angle_calculations}
		\begin{tabular}{l|c|c}
			\cellcolor{gray!10!white}\textbf{Joint name} & \cellcolor{gray!10!white}\textbf{Angle} & \cellcolor{gray!10!white}\textbf{Links for angle calculation} \\[0.1cm]\hline
            \cellcolor{gray!10!white}Right elbow (RE) & $\theta^H_1$ & $\vec{l}_{1,2}$, $\vec{l}_{2,3}$ \\[0.05cm]\hline
            \cellcolor{gray!10!white}Right shoulder roll (RSR) & $\theta^H_2$ & $\vec{l}_{2,3}$, $\vec{l}_{3,4}$ \\[0.05cm]\hline
            \cellcolor{gray!10!white}Right shoulder pitch (RSP) & $\theta^H_3$ & $\vec{l}_{{2,3}_{xy}}$, $\hat{\vec{x}}$ \\[0.05cm]\hline
            \cellcolor{gray!10!white}Head pitch (HP) & $\theta^H_4$ & $\vec{l}_{4,5}$, $\vec{l}_{5,6}$ \\[0.05cm]\hline
            \cellcolor{gray!10!white}Left shoulder roll (LSR) & $\theta^H_5$ & $\vec{l}_{4,7}$, $\vec{l}_{7,8}$ \\[0.05cm]\hline
            \cellcolor{gray!10!white}Left shoulder pitch (LSP) & $\theta^H_6$ & $\vec{l}_{{7,8}_{xy}}$, $\hat{\vec{x}}$ \\[0.05cm]\hline
            \cellcolor{gray!10!white}Left elbow (LE) & $\theta^H_7$ & $\vec{l}_{7,8}$, $\vec{l}_{8,9}$
		\end{tabular}
	\end{table}
    The joints are illustrated in Fig. \ref{fig:skeleton_annotation} for one frame of a motion demonstration.
    Here, it should be noted that the shoulders have two DOF, namely a roll and a pitch joint.
    The orientation of a shoulder's roll joint is calculated as the angle between the links connecting the shoulder with the elbow and collar; on the other hand, the orientation of the pitch joint is calculated as the angle made by the projection over the $xy$-plane of the link connecting the shoulder with the elbow (denoted as $\vec{l}_{{2,3}_{xy}}$ and $\vec{l}_{{7,8}_{xy}}$ for the right and left shoulder, respectively) with the $x$-axis of a right-handed coordinate frame as illustrated in Fig. \ref{fig:skeleton_annotation}.
    The neck has two DOF as well, namely a pitch and a yaw joint, but we do not consider motions that include yaw motion for simplicity.

    Given the angles $\theta^H_j$ of the human joints, we need to map these to the joints of the robot.
    For this, it should be noted that the configuration of QTrobot allows for a one-to-one mapping between the human and robot joints that we consider in this work; thus, the joint angles $\theta^R_j, 1 \leq j \leq n$ of the robot can be found by converting the human joint angles to the robot's joint ranges:
    \begin{equation}
        \theta^R_j = \min\theta^R_j + \frac{\max\theta^R_j - \min\theta^R_j}{\max\theta^H_j - \min\theta^H_j}\left( \theta^H_j - \min\theta^H_j \right)
        \label{eq:human_to_robot_joint_angle_conversion}
    \end{equation}
    where $\min\theta^H_j$ and $\max\theta^H_j$ are the lower and upper joint limits of the human's $j$-th joint, while $\min\theta^R_j$ and $\max\theta^R_j$ are the lower and upper limits of the robot's $j$-th joint.
    The lower and upper joint limits of the human and robot joint angles that are used in Eq. \ref{eq:human_to_robot_joint_angle_conversion} are provided in Tab. \ref{tab:human_angle_limits} and Tab. \ref{tab:qt_angle_limits}, respectively.\footnote{In practice, due to slight deviations from the values provided in the robot's specification, the actual joint limits may need to identified manually by moving the joints to their limits and reading out the measured values.}
    \begin{table}[t]
        \centering
		\caption{Human joint limits (based on \cite{Moromizato2016})}
		\label{tab:human_angle_limits}
		\begin{tabular}{l|c|c}
			\hline
			\cellcolor{gray!10!white}\textbf{Joint name} & \cellcolor{gray!10!white}\textbf{Lower limit [deg]} & \cellcolor{gray!10!white}\textbf{Upper limit [deg] } \\\hline
			\cellcolor{gray!10!white}Head pitch           &   -70 &    85 \\\hline
			\cellcolor{gray!10!white}Left elbow           &   4.3 & 142.6 \\\hline
			\cellcolor{gray!10!white}Left shoulder pitch  & -66.5 & 176.4 \\\hline
			\cellcolor{gray!10!white}Left shoulder roll   &     0 & 179.7 \\\hline
			\cellcolor{gray!10!white}Right elbow          &   4.3 & 142.6 \\\hline
			\cellcolor{gray!10!white}Right shoulder pitch & -66.5 & 176.4 \\\hline
			\cellcolor{gray!10!white}Right shoulder roll  &     0 & 179.7 \\\hline
		\end{tabular}
	\end{table}
	\begin{table}[t]
        \centering
		\caption{QTrobot's joint limits}
		\label{tab:qt_angle_limits}
		\begin{tabular}{l|c|c}
			\hline
			\cellcolor{gray!10!white}\textbf{Joint name}  & \cellcolor{gray!10!white}\textbf{Lower limit [deg]} & \cellcolor{gray!10!white}\textbf{Upper limit [deg] } \\\hline
			\cellcolor{gray!10!white}Head pitch           &  -15.3 & 21.1 \\\hline
			\cellcolor{gray!10!white}Left elbow           &   -8   & -80  \\\hline
			\cellcolor{gray!10!white}Left shoulder pitch  & -140 & 140    \\\hline
			\cellcolor{gray!10!white}Left shoulder roll   &  -21.1 & -81.1\\\hline
			\cellcolor{gray!10!white}Right elbow          &   -8.3 & -77.2\\\hline
			\cellcolor{gray!10!white}Right shoulder pitch &  140 & -140   \\\hline
			\cellcolor{gray!10!white}Right shoulder roll  &  -20 & -80    \\\hline
		\end{tabular}
	\end{table}

    The conversion results in a final representation $\Theta^R_a$ of a demonstration of $a$ in terms of the robot's joint angles, which can be executed by the robot:
    \begin{equation}
        \Theta^R_a = \begin{pmatrix}
            \vec{\theta}^R_1, \hdots, \vec{\theta}^R_T
        \end{pmatrix} =
        \begin{pmatrix}
            \theta^R_{1_1} &        & \theta^R_{T_1} \\
            \vdots         & \hdots & \vdots \\
            \theta^R_{1_n} &        & \theta^R_{T_n} \\
        \end{pmatrix}
    \end{equation}

    \subsection{Executing Demonstrated Motions on a Robot}
    \label{sec:learning:execution}

    In order to accurately imitate demonstrated movements, we use a closed-loop control system, such that the feedback is the error between the desired and the recorded angle of each robot joint.
    We use a proportional-integral-derivative (PID) controller for this purpose, which takes into account the previous, current and predicted error into consideration when calculating the control output.
    The error between a desired position $\theta^R_{t_j}$ and the measured position $\hat{\theta}^R_{t_j}$ for a joint $j, 1 \leq j \leq n$ is
    \begin{equation}
        e^j_t = \theta^R_{t_j} - \hat{\theta}^R_{t_j}
    \end{equation}
    Given $e^j_t$, the control output $u^j$ for joint $j$ is found as
    \begin{equation}
        \begin{split}
            u^j &= K_p \left(e^j_t + \frac{1}{T_{i}} \int_{0}^{t} e^j_\tau d\tau + T_{d} \frac{de^j_t}{dt}\right) \\
            &= K_p e^j_t + K_i\int_{0}^{t} e^j_\tau d\tau + K_d\frac{de^j_t}{dt}
        \end{split}
        \label{eq:pid}
    \end{equation}
    where $K_p$, $K_i$, and $K_d$ are controller constants.
    Eq. \ref{eq:pid} describes a continuous time expression \cite{Visioli2006}, where the setpoint $\theta^R_{t_j}$ for each joint $j$ is the angle of $j$ in $\Theta^R_a$.
    The control strategy, summarised in Alg. \ref{alg:demonstration_reproduction}, is performed continuously for all joints and time points in $\Theta^R_a$.

    \begin{algorithm}[tp]
        \caption{Robot execution of a demonstrated trajectory. Here, $\epsilon$ is an error tolerance, $\Delta t$ is a control time step, and \texttt{getJointAngle} is a function that returns the measured value of a given joint.}
        \label{alg:demonstration_reproduction}
        \begin{algorithmic}[1]
            \Function{reproduceMotion}{$\Theta^R_a$, $T$, $n$, $\Delta t$}
                \For{$t$ \assign $1$ \algto $T$}
                    \State $e^j_t \leftarrow \infty, \; \forall j \in [1,n]$
                    \State $e^j_{int} \leftarrow 0, \; \forall j \in [1,n]$
                    \For{$j$ \assign $1$ \algto $n$}
                        \While{$e^j_t \geq \epsilon$}
                            \State $\hat{\theta}^R_{t,j}$ \assign \texttt{getJointAngle}($j$)
                            \State $e^j_t$ \assign $\theta^R_{t,j} - \hat{\theta}^R_{t,j}$
                            \State $e^j_{dot}$ \assign $\frac{e^j_t - e^j_{t-1}}{\Delta t}$
                            \State $e^j_{int}$ \assign $e^j_{int} + e^j_t\Delta t$
                            \State $u^j$ \assign $\hat{\theta}^R_{t,j} + (K_p e^j_t + K_i e^j_{int} + K_d {e^j_{dot}})$
                        \EndWhile
                    \EndFor
                \EndFor
            \EndFunction
        \end{algorithmic}
    \end{algorithm}

    Let $\overline{\theta}^R_{t,j}$ be the achieved angle for joint $j$ after executing the above control law given $\theta^R_{t,j}$ as a target joint angle.
    To evaluate the reproduction quality of a complete motion trajectory, we calculate the average error over all joints $j$ for each target $\vec{\theta}^R_t, 1 \leq t \leq T$; in other words, the average joint error for target $\vec{\theta}^R_t$ is
    \begin{equation}
        E_t = \frac{1}{n}\sum_{j=1}^{n}\theta^R_{t,j} - \overline{\theta}^R_{t,j}
        \label{eq:avg_tracking_error}
    \end{equation}

%% file: subfiles/experiments.tex
    \section{EXPERIMENTS}
    \label{sec:experiments}

    We evaluate the described method by letting QTrobot learn and execute upper-body dance moves, each of which is demonstrated by multiple experimental participants.
    We particularly evaluate the reproduction quality over different users, namely we execute motions as demonstrated by the participants so that the quality over participants and moves can be evaluated separately.

    \subsection{Experimental Setup}
    \label{sec:experiments:setup}

    To collect learning data, we performed an experiment in which QTrobot was placed on a table; the participants were standing between $1.5$--$2m$ from the table as shown in Fig. \ref{fig:skeleton_annotation} so that the robot could see the upper body at all times.
    The robot showed 12 pre-programmed upper-body dance moves to each participant; the participant then reproduced the robot's motions, such that the objective is to let the robot perform these motions based on the demonstrations.\footnote{The motions performed by the robot can be seen at \url{https://youtu.be/McpLujg5gR8}.}
    We performed the experiment with 20 adult participants; each of them demonstrated the 12 moves in a sequence, namely the robot first showed each move and then waited 4 seconds for the person to execute the motion before proceeding to the next move.
    The participants in this experiment were master's students and two PhD students, most of which have some experience with robots, but only a few of them had prior interaction with QTrobot; four of the participants in the experiments are part of the project team and work with the robot on a regular basis.
    The demonstrations of the 12 moves were then manually segmented from the skeleton data.\footnote{We use the built-in Nuitrack for extracting skeleton data on QTrobot: \url{https://nuitrack.com}}
    To evaluate the execution on the robot, we executed the moves based on the demonstrations by the individual participants and measured the reproduction quality as described in the previous section.\footnote{Multiple demonstrations of the same motion will allow us to learn a more versatile model of each move, namely such a model could potentially be used to sample motions from a trajectory envelope encompassing all demonstrations, similar to \cite{haidu2015}.}
    We only present a subset of these results in this paper in the interest of space.

    \subsection{Results}
    \label{sec:experiments:results}

\begin{figure}[t]
        \begin{subfigure}{0.475\linewidth}
            \centering
            \includegraphics[width=\linewidth]{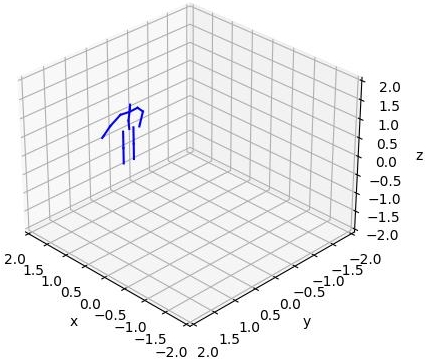}
        \end{subfigure}
        \begin{subfigure}{0.475\linewidth}
            \centering
            \includegraphics[width=\linewidth]{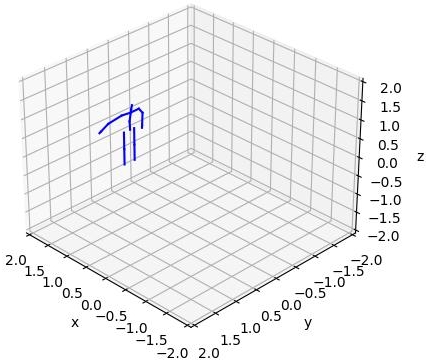}
        \end{subfigure}
        \newline
        \begin{subfigure}{0.475\linewidth}
            \centering
            \includegraphics[width=\linewidth]{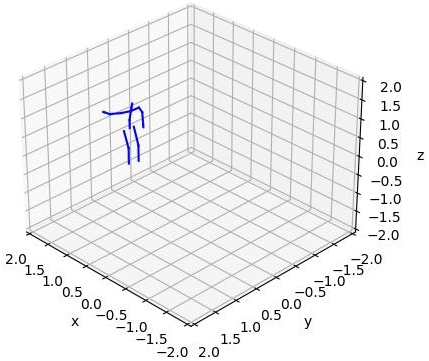}
        \end{subfigure}
        \begin{subfigure}{0.475\linewidth}
            \centering
            \includegraphics[width=\linewidth]{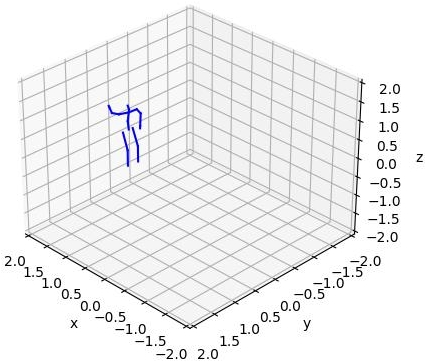}
        \end{subfigure}
        \caption{Illustration of multiple (non-consecutive) captured skeleton frames from one action (teapot) for one participant. All positions are shown in meters. Noisy frames appear in the data by sudden position jumps of one or more joints to non-desired configurations; these are removed before learning.}
        \label{fig:skeleton_frames}
    \end{figure}

    Fig. \ref{fig:skeleton_frames} illustrates a few frames of one of the actions performed by one of the participants; this particular motion is referred to as \emph{teapot} and involves putting one of the hands on the hip, while the other hand is to the side and raised high, slightly above the head's level.
    For a few of the participants and motions, some of the recorded skeleton frames were noisy; noisy frames manifest themselves by the position of one or more joints exhibiting a sudden jump to a value that is significantly different from the value in the previous recorded frame.
    Such frames were removed from the learning data during the manual action segmentation.

\begin{figure}[t]
        \begin{subfigure}{0.495\linewidth}
            \centering
            \includegraphics[width=\linewidth]{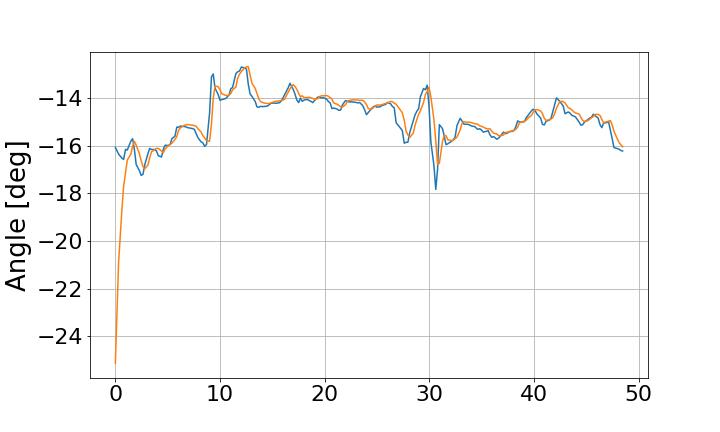}
            \caption{Teacup}
        \end{subfigure}
        \begin{subfigure}{0.495\linewidth}
            \centering
            \includegraphics[width=\linewidth]{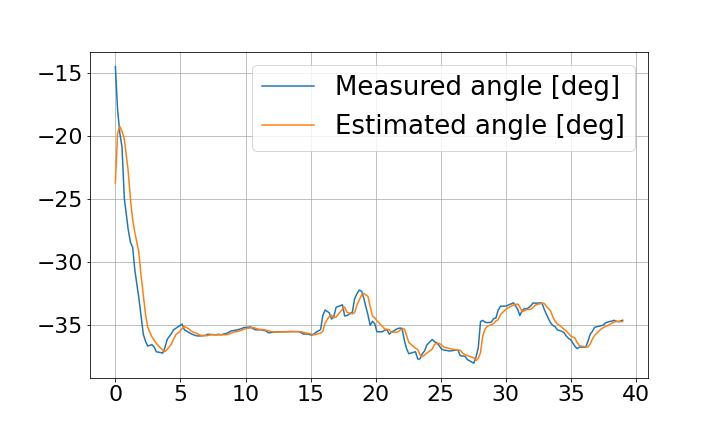}
            \caption{Ladle}
        \end{subfigure}
        \newline
        \begin{subfigure}{0.495\linewidth}
            \centering
            \includegraphics[width=\linewidth]{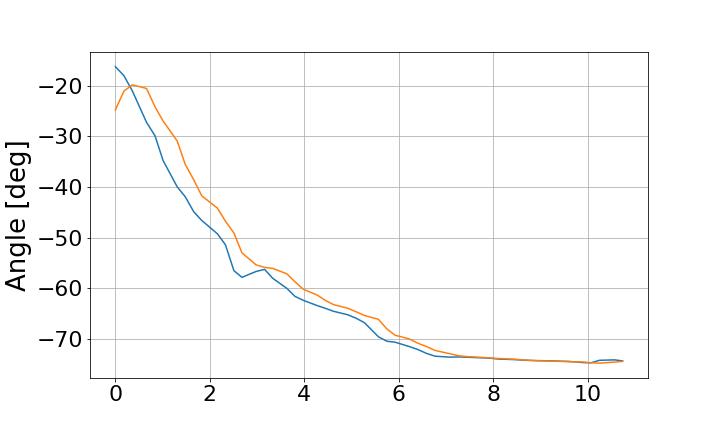}
            \caption{Fork}
        \end{subfigure}
        \begin{subfigure}{0.495\linewidth}
            \centering
            \includegraphics[width=\linewidth]{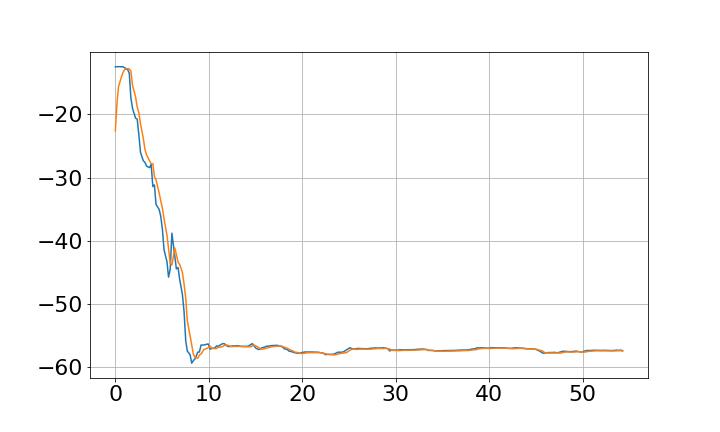}
            \caption{Spoon}
        \end{subfigure}
        \newline
        \begin{subfigure}{0.495\linewidth}
            \centering
            \includegraphics[width=\linewidth]{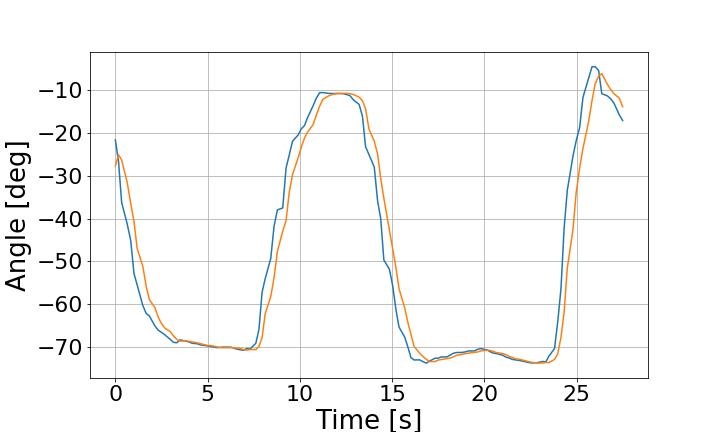}
            \caption{Knife}
        \end{subfigure}
        \begin{subfigure}{0.495\linewidth}
            \centering
            \includegraphics[width=\linewidth]{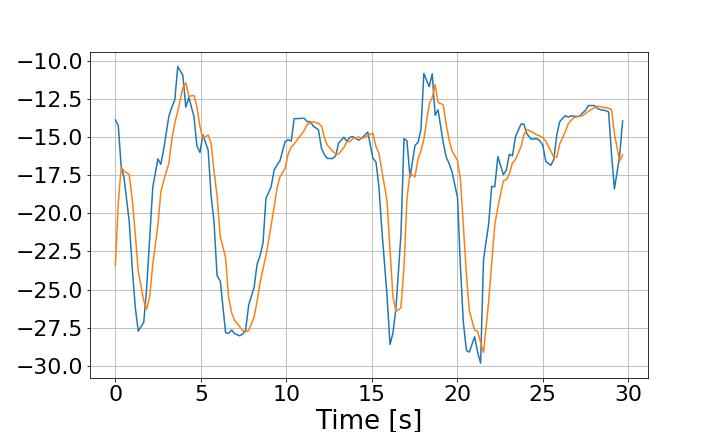}
            \caption{Salt shaker}
        \end{subfigure}
        \caption{Angle tracking error of the PID controller for the right elbow joint of one participant for six different actions. Here, \emph{measured} represents a target angle, while \emph{estimated} is the angle computed by the controller.}
        \label{fig:controller_errors}
    \end{figure}

    To tune the parameters of the controller described in section \ref{sec:learning:execution} and investigate its tracking quality independently of the execution on the robot, we perform an evaluation of the algorithm in which we attempt trajectory tracking in an open loop, without measuring the robot's joint angles.
    For this, the initial joint angles (before tracking) are always set to the same starting position and the controller is then executed as in Alg. \ref{alg:demonstration_reproduction}, with the exception that the joint angles are not measured, but the control outputs are instead assumed to represent the measurements.
    The results of this evaluation are presented in Fig. \ref{fig:controller_errors} for the right elbow joint of one of the participants over six different actions.
    As these results illustrate, after tuning the parameters, the angle trajectories can be tracked fairly accurately for different actions, including periodic motions such as in the case of the \emph{salt shaker} action.
    This does not hold for all actions and participants, as there are some participants for which the tracking error is relatively large (for instance, more than $10^o$).
    We believe that this is due to the fact that a few of the demonstrations were noisy, which meant that the density of actual frames that represent the desired motion was lower than in motions with low demonstration noise.

\begin{figure}[t]
        \begin{subfigure}{\linewidth}
            \centering
            \includegraphics[width=\linewidth]{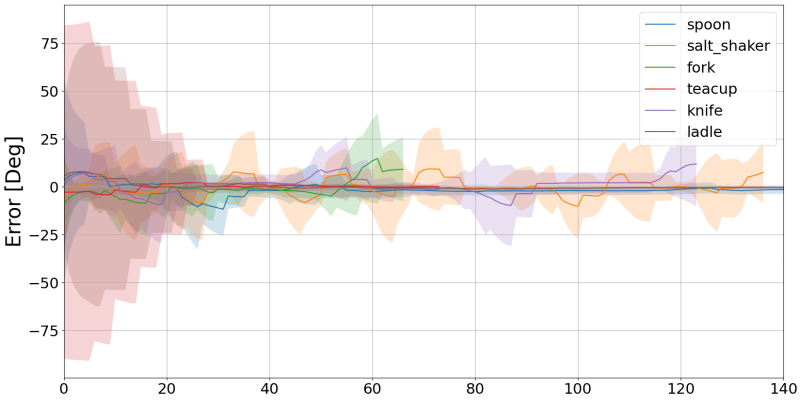}
        \end{subfigure}
        \newline
        \begin{subfigure}{\linewidth}
            \centering
            \includegraphics[width=\linewidth]{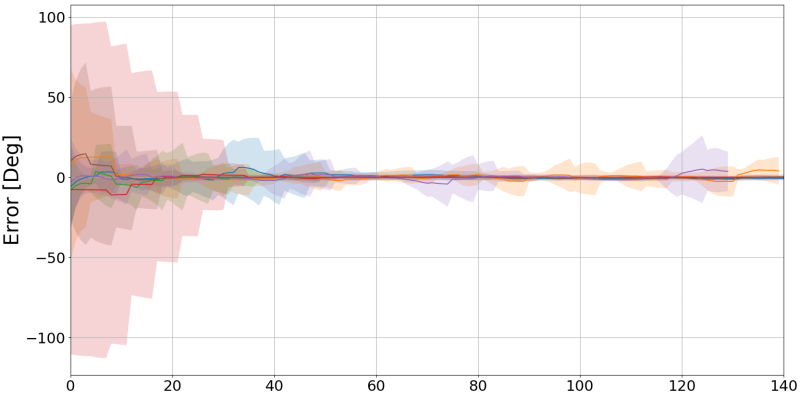}
        \end{subfigure}
        \newline
        \begin{subfigure}{\linewidth}
            \centering
            \includegraphics[width=\linewidth]{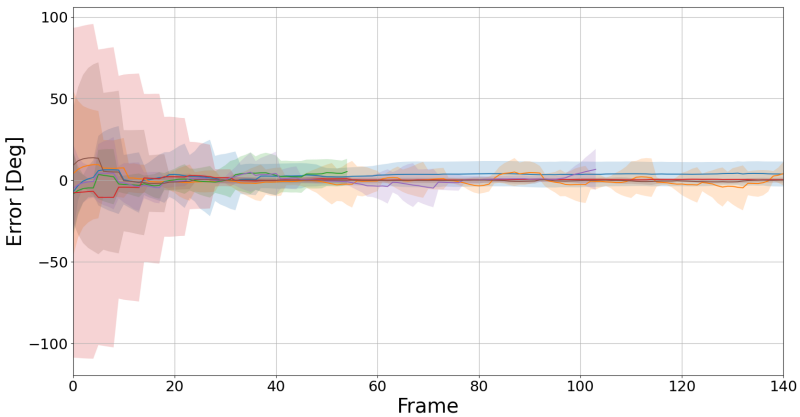}
        \end{subfigure}
        \caption{Average trajectory execution errors for three participants and six different actions. The shaded regions represent the standard deviation of the error. The deviation is large in the first few tracking frames because we start the imitation at an arbitrary initial robot configuration, but decreases considerably as the robot starts tracking the trajectory closely.}
        \label{fig:trajectory_execution_error}
    \end{figure}

    Finally, we evaluate the trajectory execution on the robot, namely we execute Alg. \ref{alg:demonstration_reproduction} for different participants and actions and calculate the average reproduction error as in Eq. \ref{eq:avg_tracking_error}.
    Fig. \ref{fig:trajectory_execution_error} shows the errors for three participants on the same actions as in Fig. \ref{fig:controller_errors}.
    When performing this evaluation, the actions were executed in a sequence; thus, the error in the first tracking frames has a large standard deviation, as some of the joints were far from their desired configuration at the beginning of each trajectory.
    As the trajectory was tracked for a longer period, both the error and the standard deviation decreased noticeably, which suggests that the robot executed the motion rather accurately.
    The plots also illustrate that the error increases at the turning points of the periodic motions, but then stabilises reasonably quickly.

%% file: subfiles/conclusions.tex
    \section{DISCUSSION AND CONCLUSIONS}
    \label{sec:discussion}

    This paper presented a method for acquiring motions from human demonstrations and reproducing them on a robot, with a particular focus on upper-body motions applied to QTrobot.
    Using observed human joint positions, which are represented as skeletons collected by an RGB-D camera on the robot, we calculate the joint angles and directly map those to the robot's joints; the execution is then performed by a PID position controller, which minimises the joint angle tracking error.
    We evaluated the method in an experiment in which different participants demonstrated a set of upper-body dance moves to the robot, which were then executed by the robot.
    The results, in particular the average error over all joints for a complete trajectory, suggest that the method is feasible to be used for motion demonstration to the robot; however, the execution error is affected by noisy frames in the recorded skeleton demonstrations.

    There are various aspects in which the proposed approach could be extended.
    As mentioned in section \ref{sec:related_work}, we need to perform experiments in which the volumetric-based self-collision avoidance approach \cite{Corrales2011} is used to protect the robot during execution.
    For the experiments in this paper, we performed manual segmentation of the demonstrated motions and manual noise removal; we would particularly like to automate the noise removal by using change detection for identifying noisy frames.
    During the data collection in our experiments, we noticed that there was a difference in how participants interpreted the moves that were originally shown by the robot: some participants chose to mirror the robot, while others took the robot's perspective when demonstrating the actions; this difference in perspective-taking is an aspect that we believe is particularly important to study in the context of ASD.
    We would also like to perform experiments in which motions are demonstrated by non-expert users (such as therapists) and integrate the demonstrated motions in activities that are used in therapeutic activities.
    Related to this, the proposed approach should be integrated with an action recognition method, which would make it possible to apply the acquired trajectories in activities where the robot performs the trajectories and a child has to repeat them.